\documentclass[11pt]{article}
\PassOptionsToPackage{numbers}{natbib} % to support numeric citations for space reduction
\usepackage[preprint]{neurips_2026}
\usepackage[utf8]{inputenc}
\usepackage[T1]{fontenc}
\usepackage{hyperref}
\usepackage{url}
\usepackage{booktabs}
\usepackage{amsfonts}
\usepackage{amsmath}
\usepackage{nicefrac}
\usepackage[expansion=false]{microtype}
\usepackage{graphicx}
\usepackage{xcolor}
\usepackage{multirow}
\usepackage{tabularx}
\usepackage{enumitem}
\usepackage{float}
\usepackage{listings}
\definecolor{bestblue}{RGB}{0,100,200}
\definecolor{bestgreen}{RGB}{0,150,50}
\definecolor{collapsered}{RGB}{200,0,0}
\definecolor{phgray}{RGB}{130,130,130}
\newcommand{\best}[1]{\textcolor{bestblue}{\textbf{#1}}}
\newcommand{\bestp}[1]{\textcolor{bestgreen}{\textbf{#1}}}
\newcommand{\collapse}[1]{\textcolor{collapsered}{#1}}

\newcommand{\care}{Ca\textsc{re}}

% ...
\title{%
  \raisebox{-0.15\height}{\includegraphics[height=0.9em]{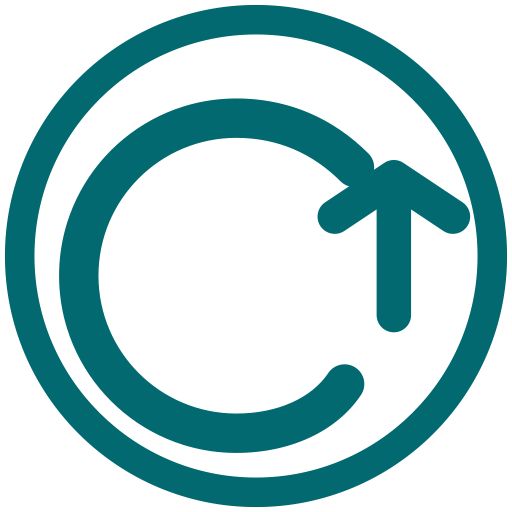}}%
  \hspace{0.3em}%
  \care: Compute-aware Remasking Evaluation\\Protocol for Masked Diffusion Language Models
}

\author{%
  Yash Shah$^{1}$\thanks{Equal contribution.} \\
  Arizona State University \\
  \texttt{yshah124@asu.edu}
  \And
  Abhijit Chakraborty$^{2}$\footnotemark[1] \\
  MongoDB \\
  \texttt{abhijit.chakraborty@mongodb.com}
  \And
  Vivek Gupta$^{1}$ \\
  Arizona State University \\
  \texttt{vgupt140@asu.edu}
}
\begin{document}
\maketitle

% -----------------------------------------------------------------------

\begin{abstract}
Masked diffusion language models are outpacing the evaluation standards
needed to reliably interpret their progress. MDLMs are now competitive
with autoregressive LMs, yet seven recent remasking papers evaluate under
incompatible settings, rendering their strategy rankings largely
incomparable. Existing studies vary nominal step counts, metrics, and
sampling temperatures, but no work jointly controls these factors, leaving
open whether reported gains reflect algorithms or evaluation artifacts.
We present \care{}\footnote{https://anonymous.4open.science/w/CaRE-846C/, We release the protocol, implementation, and leaderboard to ensure future remasking claims are reproducible and robust.}, a compute-aware evaluation framework for auditing masked-discrete MDLM remasking by standardizing actual NFE, multi-metric reporting, and stochasticity. We release the protocol, implementation, and leaderboard to ensure future remasking claims are reproducible and robust. Applied to 7 remasking strategies across
LLaDA-8B-Base and Dream-7B-Base, 4 stochasticity levels, and 3 step
budgets on OpenWebText and LM1B, \care{} shows that
(i)~temperature explains the majority of MAUVE variance ($\eta^2{=}0.91$),
(ii)~compute-matched comparisons reverse several published strategy
rankings, and (iii)~informed remasking and stochastic unmasking are in
tension, with high-entropy remasking reducing MAUVE by $0.296$ at 256
steps and \texttt{unmask\_temp}${=}0.25$ ($p{=}0.020$). A \care{}
leaderboard covers 12 open-weight MDLMs (150M--8B parameters), where
the interaction direction holds across architectures and scales. These
results reveal that current MDLM evaluations can systematically conflate
algorithmic improvements with hidden choices of compute and stochasticity,
and \care{} releases a seven-point protocol and implementation to make
future remasking claims reproducible.
\end{abstract}
 
% -----------------------------------------------------------------------
\section{Introduction}
\label{sec:intro}
% -----------------------------------------------------------------------
 
For over a decade, autoregressive language models have treated left-to-right
generation less as a design choice than as a default. Masked diffusion language
models (MDLMs) challenge this assumption by reformulating generation as
iterative parallel denoising over the full sequence. The emergence of
large-scale MDLMs such as LLaDA-8B~\cite{nie2025large} and
Dream-7B~\cite{ye2025dream} shows that this paradigm has moved from
theoretical curiosity to a competitive alternative to autoregressive LMs.
\textbf{Throughout this paper, ``remasking'' refers to mask re-introduction
during denoising in \emph{masked discrete-token} diffusion language models;
continuous-state and flow-based DLMs use a different inference paradigm and
are out of scope (\S\ref{sec:limitations}).}
\begin{figure}[H]
\centering
\includegraphics[width=\linewidth]{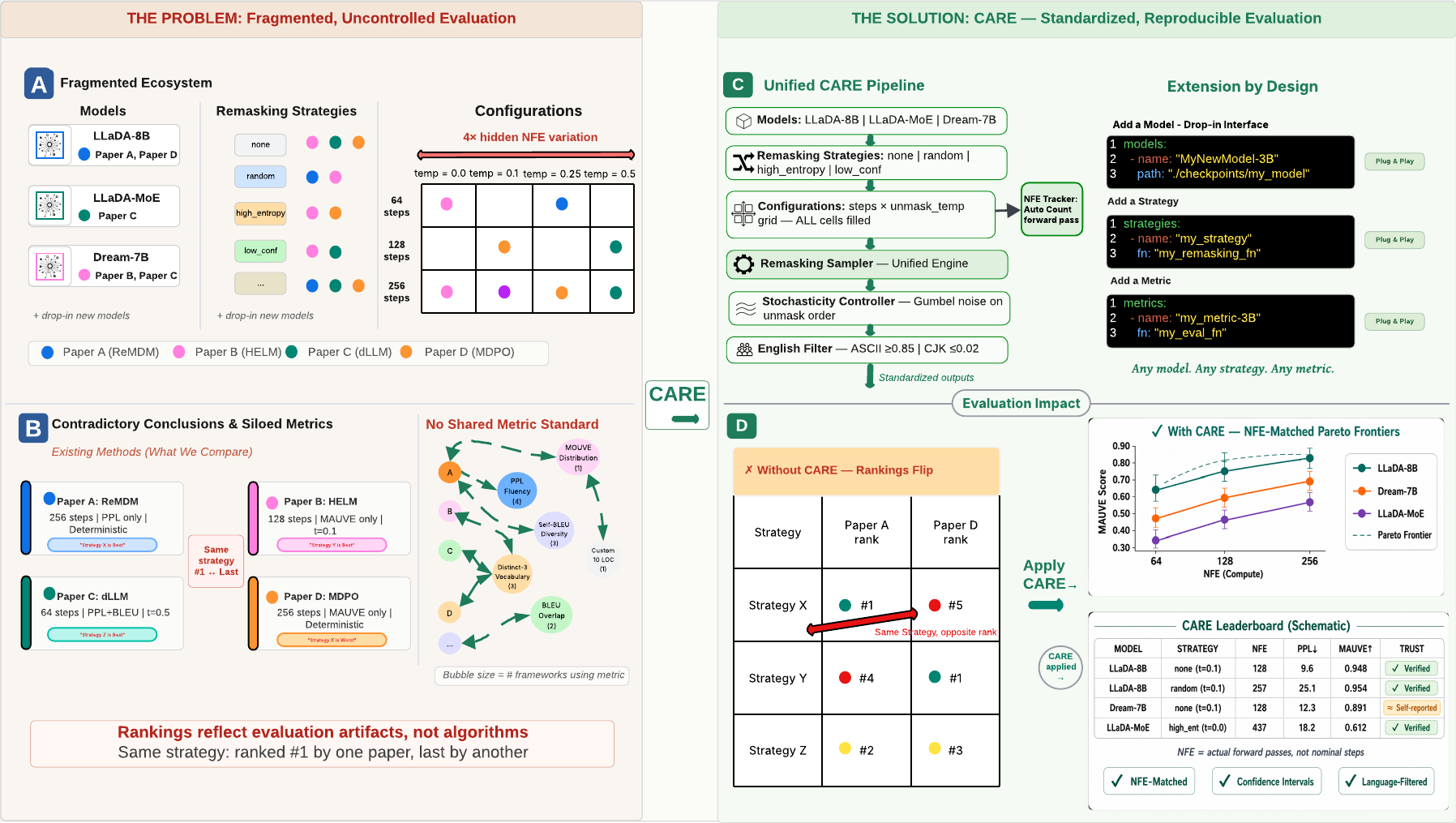}
\vspace{-3mm}
\caption{\small The evaluation gap: seven remasking papers evaluate under different conditions, producing contradictory rankings. Three confounds (compute, metric, stochasticity) each independently reverse conclusions. Controlling all three jointly reveals their interaction ($p{=}0.020$; \texttt{none} vs.\ \texttt{high\_entropy}, LLaDA-8B-Base, OWT, 256 steps, $t{=}0.25$, 3 seeds, paired $t$-test).}
\label{fig:problem}
\end{figure}

\paragraph{The evaluation crisis.}
In the last six months, seven remasking strategies~\cite{wang2025remasking,
zhai2026core,kim2025fine,he2025mdpo,huang2025don,asano2026unmask,
kim2026stop} have been proposed to improve MDLM inference by selectively
re-introducing masks during denoising. Each is evaluated under different step
budgets, metrics, and sampling temperatures, producing contradictory rankings
where the same strategy is reported as best in one paper and worst in another
(Figure~\ref{fig:problem}). The image generation community faced a similar
situation before Clean-FID~\cite{parmar2022aliased} standardized evaluation, and
machine translation before SacreBLEU~\cite{post2018call} enforced tokenizer
consistency. In both cases, part of the apparent progress was later attributed
to evaluation inconsistency rather than algorithmic improvement. MDLMs are now
at a similar inflection point.
\newline\textbf{The hidden interaction.}
Individual confounds are documented in prior work: PPL--MAUVE
disagreement~\cite{pillutla2021mauve}, temperature
sensitivity~\cite{chang2022maskgit,hayakawa2025demystifying}, NFE
formalization~\cite{ou2024your}, and decoding-hyperparameter
dominance~\cite{Holtzman2019TheCC}. The community has flagged the
problem~\cite{peng2025efficient,zhou2026dllm,yang2026dare,pynadath2026generative} but
without joint control. \care{} combines these strands and reveals that
\emph{informed remasking and stochastic unmasking are in consistent tension}.
Stochastic unmasking is the dominant diversity lever, improving MAUVE from
$0.66$ to $0.98$ with a single temperature change---a gain of $0.32$ that
exceeds the maximum gap between any two strategies at the same temperature.
Yet at higher compute budgets, high-entropy remasking collapses this benefit
by $0.296$ MAUVE points ($p{=}0.020$) through persistent token churn that
narrows distributional coverage while preserving surface-level diversity. To
our knowledge, no concurrent MDLM evaluation surfaces this interaction,
because detecting it requires controlling all three confounds simultaneously.
\newline\textbf{Contributions.}
\textbf{(1)~Interaction finding}: informed remasking and stochastic unmasking
are in consistent tension ($p{=}0.020$); a three-way ANOVA shows temperature
dominates ($\eta^2{=}0.91$) with significant strategy~$\times$~temperature
interaction ($\eta^2{=}0.47$). Unlike prior frameworks that control at most
two confounds (Table~\ref{tab:frameworks}), detecting this interaction
requires compute, metric, and stochasticity to be controlled \emph{jointly},
which is why it has not appeared in earlier MDLM evaluation work.
\textbf{(2)~Compute-matched reversal}: strategy
rankings flip when actual NFE is equated, on both unconditional generation
and HumanEval. \textbf{(3)~Seven-point protocol and leaderboard}: a
reproducible standard validated on OpenWebText and LM1B, with a public
leaderboard over twelve open-weight MDLMs. A frontier overlay
(\S\ref{sec:frontier}) shows the interaction is invisible in PPL-entropy
space, positioning \care{} as complementary to Generative
Frontiers~\cite{pynadath2026generative}.
 
% -----------------------------------------------------------------------
\section{The \care~ Framework and Protocol}
\label{sec:framework}
% -----------------------------------------------------------------------
 
\care~ provides a unified \texttt{RemaskingSampler} with four design
principles: (1)~automatic NFE tracking per generation; (2)~four default
metrics (PPL via Llama-3-8B, MAUVE, Self-BLEU, Distinct-3);
(3)~\texttt{unmask\_temp} as a required parameter with sweep support;
(4)~character-level English filter (ASCII~$\geq$~0.85, CJK~$\leq$~0.02).
Adding a model takes $\sim$50 LOC; adding a strategy takes $\sim$20 LOC.
Appendix~\ref{app:customization} gives the complete \texttt{HighEntropy}
implementation as shipped in the released codebase, showing that a full
remasking strategy fits in roughly 20 lines.
Prompt formats for unconditional generation, HumanEval, and GSM8K follow
standard \texttt{lm-eval} configurations; representative templates appear
in Appendix~\ref{app:prompts}.
The full experimental scope spans 12 open-weight MDLMs (150M--8B parameters),
5 evaluation settings, and 7 remasking strategies; details appear in
Section~\ref{sec:validation}.

\begin{table}[t]
\centering\small
\caption{Seven-point \care~ protocol. Each omission independently reverses
published conclusions.}
\label{tab:protocol}
\vspace{-2mm}
\begin{tabularx}{\textwidth}{@{}p{3.1cm}X@{}}
\toprule
\textbf{Practice} & \textbf{What goes wrong without it} \\
\midrule
1.\ Report actual NFE        & ``256 steps'' hides 128--513 forward passes ($4\times$ range) \\
2.\ Multi-metric              & PPL and MAUVE rankings reverse (Tables~\ref{tab:det_llada},~\ref{tab:mauve_temps}) \\
3.\ External-evaluator PPL    & Shared vocabulary inflates scores; we use Llama-3-8B \\
4.\ Control stochasticity     & $t{:}0.0{\to}0.1$ shifts MAUVE by $0.3{-}0.5$ (model-dependent; ${\approx}0.3$ on LLaDA-8B-Base, ${\approx}0.5$ on LLaDA-MoE), larger than any strategy gap \\
5.\ Language filtering        & Off-target language contaminates MAUVE unpredictably \\
6.\ Include \texttt{none} baseline & Improvements may be vs.\ artificially weak reference \\
7.\ Report uncertainty        & Single-seed MAUVE margins $<$0.1 unreliable (SE up to 0.05) \\
\bottomrule
\end{tabularx}
\end{table}
 
% -----------------------------------------------------------------------
\section{Related Work}
\label{sec:related}
% -----------------------------------------------------------------------
 
\paragraph{Remasking strategies.}
ReMDM~\cite{wang2025remasking} identifies PPL hacking and advocates MAUVE, but
applies nucleus sampling as an on/off toggle without isolating stochasticity.
CoRe~\cite{zhai2026core} uses compute-matched baselines for a single
strategy; Prism~\cite{bai2026prism} scales test-time compute via hierarchical NFE; UnMaskFork~\cite{misaki2026unmaskfork}
finds temperature-based scaling degrades MDLMs. \care~ codifies these
ad-hoc practices and reveals the interaction they cannot detect.
\newline\textbf{What \care{} adds over existing frameworks.}
Existing frameworks each control at most two of the three confounds
\care{} identifies. \textbf{HELM}~\cite{liang2022holistic} is multi-metric
but targets autoregressive LLMs with no notion of NFE or remasking.
\textbf{Generative Frontiers}~\cite{pynadath2026generative} treats
temperature as a curve parameter and decomposes generation into
PPL--entropy space, but does not compare remasking strategies and
leaves NFE implicit (Section~\ref{sec:frontier}).
\textbf{dLLM}~\cite{zhou2026dllm} flags hyperparameter sensitivity but
provides no remasking strategy-comparison protocol.
\textbf{ReMDM}~\cite{wang2025remasking} controls NFE and advocates MAUVE
but conflates stochasticity with nucleus sampling on/off, leaving
rankings entangled with the variable \care{} identifies as dominant
($\eta^2{=}0.91$).
\care{} is, to our knowledge, the first framework to jointly control
compute, metric, and stochasticity in a single runnable protocol
(Table~\ref{tab:frameworks}) and to expose their interaction as a
measurable object; DARE~\cite{yang2026dare} and
d3LLM~\cite{qian2026d3llm} are complementary post-training and
parallelism frameworks.
 
\begin{table}[t]
\centering\small
\caption{Framework comparison. \care~ is the only one to jointly control all
three confounds and identify their interaction.}
\label{tab:frameworks}
\vspace{-2mm}
\setlength{\tabcolsep}{4pt}
\begin{tabular}{lccccc}
\toprule
Framework & Multi-metric & NFE & Stoch.\ ctrl & Remask & Interaction \\
\midrule
HELM~\cite{liang2022holistic}                    & \checkmark & ---     & ---     & ---     & --- \\
dLLM~\cite{zhou2026dllm}                & \checkmark & partial & partial & ---     & --- \\
Gen.\ Frontiers~\cite{pynadath2026generative} & PPL-ent.   & implicit & implicit & ---  & --- \\
ReMDM~\cite{wang2025remasking}              & \checkmark & \checkmark & partial & partial & --- \\
\textbf{\care~ (ours)}                  & \checkmark & \checkmark & \checkmark & \checkmark & \checkmark \\
\bottomrule
\end{tabular}
\end{table}
 
\par\textbf{Evaluation methodology.}
Clean-FID~\cite{parmar2022aliased} and SacreBLEU~\cite{post2018call} are \care's
direct analogs. LM Eval Harness~\cite{gao2023framework} standardizes broader LLM
evaluation; Zheng et al.~\cite{Zheng2024MaskedDM} identify fp32
truncation as implicit temperature in MDMs;
Holtzman et al.~\cite{Holtzman2019TheCC} show decoding hyperparameters dominate
AR generation.
 
% -----------------------------------------------------------------------
\section{Validation Experiments}
\label{sec:validation}
% -----------------------------------------------------------------------
 
\paragraph{Evaluation scope.}
Experiments span \textbf{12 open-weight MDLMs} (150M--8B parameters;
dense, MoE, and Soft-Masked variants) and \textbf{5 evaluation settings}:
(1)~unconditional generation on OpenWebText~\cite{Gokaslan2019OpenWeb}
and LM1B~\cite{Chelba2013OneBW} (MAUVE, PPL, Self-BLEU, Distinct-3);
(2)~code generation on HumanEval (pass@1);
(3--5)~arithmetic, commonsense, and multi-step reasoning on
GSM8K~\cite{cobbe2021gsm8k}, HellaSwag~\cite{Zellers2019HellaSwagCA},
and BBH~\cite{suzgun2023challenging}---evaluated via log-likelihood
scoring as sampling-invariant sanity checks. Additional experiments and setup has been dicussed in Appendix~\ref{app:experiment}.
Strategies cover all remasking methods supported by LLaDA and Dream
(Appendix~\ref{app:Taxonomy}).

\subsection{Confound 1: Nominal Steps $\neq$ Actual Compute}
\label{sec:c1}
 
If strategies are compared at the same nominal step count without
controlling actual NFE, any quality difference could reflect a compute
advantage rather than an algorithmic one.

At 256 nominal steps, actual NFE ranges from 128 (\texttt{none}) to 513
(\texttt{running\_conf.})---a $4\times$ difference (Table~\ref{tab:nfe}).
For \texttt{none}, which never reintroduces masks, each of the $L{=}128$
output positions is unmasked exactly once; the model therefore requires
exactly 128 forward passes regardless of any nominal step count above 128,
explaining the factor-of-two between the nominal budget (256) and actual
NFE (128). Strategies that remask already-decided tokens incur additional
forward passes proportional to their remasking rate.
 
\begin{table}[h]
\centering\small
\caption{Actual NFE at 256 nominal steps, LLaDA-8B-Base.}
\label{tab:nfe}
\vspace{-2mm}
\begin{tabular}{lcc}
\toprule
Strategy & Actual NFE & Ratio \\
\midrule
\texttt{none} & 128 & 1.0$\times$ \\
\texttt{random} & 257 & 2.0$\times$ \\
\texttt{high\_entropy / low\_conf / conf\_ent / agreement} & 437 & 3.4$\times$ \\
\texttt{running\_confidence} & 513 & 4.0$\times$ \\
\bottomrule
\end{tabular}
\end{table}
 
\textbf{Compute-matched quality.} Fixing actual NFE at $\{128, 256, 512\}$
shows that \texttt{none} dominates MAUVE at every budget, and the gap widens
with more compute (Table~\ref{tab:compute_matched}).
 
\begin{table}[h]
\centering\small
\caption{Compute-matched quality (LLaDA-8B-Base, OWT, prefix=64, gen=128, $t{=}0.0$, English-filtered; PPL via Llama-3-8B). \texttt{none} rows use 3-seed mean (seeds 1--3); \texttt{random} and \texttt{high\_entropy} rows are single-seed (seed~1). These are separate experimental runs from Table~\ref{tab:det_llada}: strategies are run at different nominal step counts to equate actual NFE across configurations.}
\label{tab:compute_matched}
\vspace{-2mm}
\begin{tabular}{rl rr r}
\toprule
NFE & Strategy & PPL$\downarrow$ & MAUVE$\uparrow$ & $\Delta$MAUVE vs.\ none \\
\midrule
\multirow{3}{*}{128}
 & \texttt{none}          & 7.6  & \textbf{0.601} & --- \\
 & \texttt{random}        & 25.9 & 0.574 & $-$0.027 \\
 & \texttt{high\_entropy} & 16.8 & 0.589 & $-$0.012 \\
\midrule
\multirow{3}{*}{256}
 & \texttt{none}          & 7.2  & \textbf{0.645} & --- \\
 & \texttt{random}        & 27.6 & 0.598 & $-$0.047 \\
 & \texttt{high\_entropy} & 18.9 & 0.583 & $-$0.062 \\
\midrule
\multirow{3}{*}{512}
 & \texttt{none}          & 7.0  & \textbf{0.671} & --- \\
 & \texttt{random}        & 29.2 & 0.612 & $-$0.059 \\
 & \texttt{high\_entropy} & 20.4 & 0.557 & $-$0.114 \\
\bottomrule
\end{tabular}
\end{table}
 
\subsection{Confound 2: PPL and MAUVE Prefer Different Strategies}
\label{sec:c2}
 
Even with compute matched, a single metric can award different winners---confirming
that metric choice is an independent confound, not a proxy for the same
underlying quality.

\begin{figure}[htbp]
\centering
\includegraphics[width=0.78\textwidth]{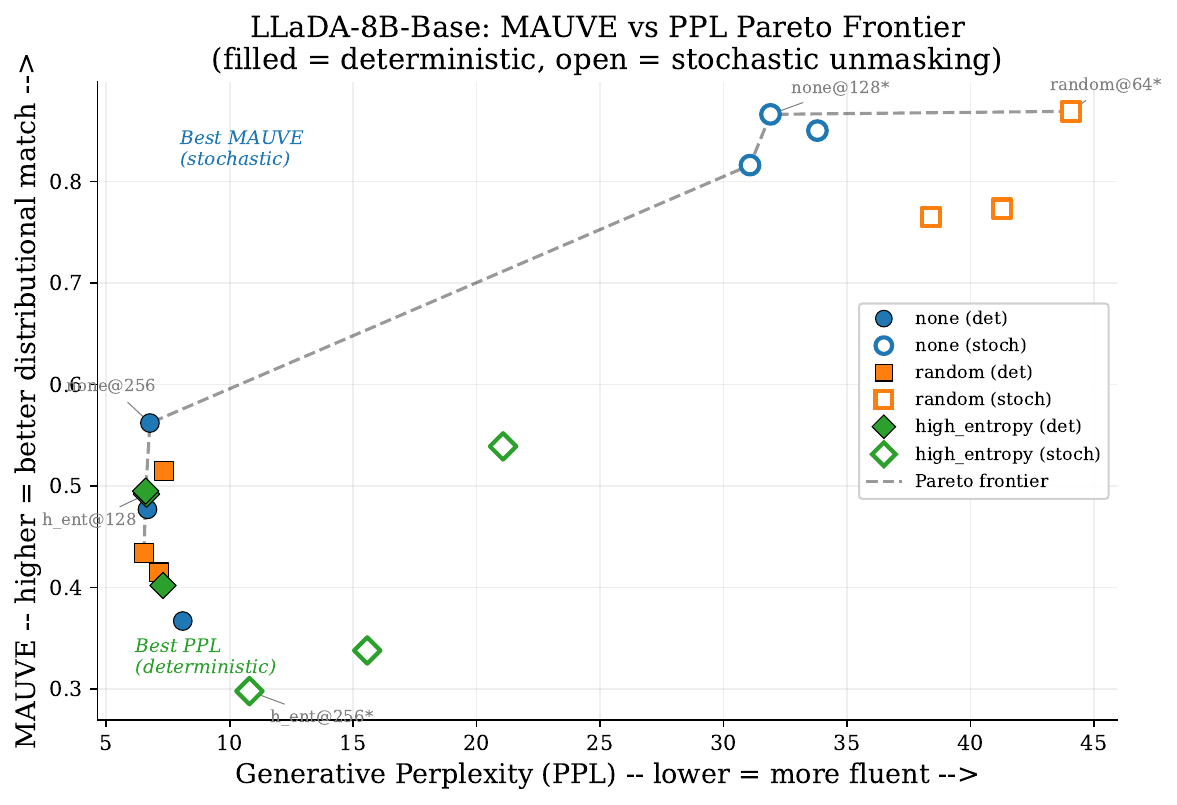}
\vspace{-3mm}
\caption{LLaDA-8B-Base, OWT. MAUVE vs PPL across step budgets (64/128/256) and temperatures. Open markers = stochastic ($t>0$); filled = deterministic ($t{=}0.0$). Stochastic configurations dominate the Pareto frontier at every PPL level.}
\label{fig:pareto}
\end{figure}

At 128 deterministic steps, \texttt{high\_entropy} achieves the best MAUVE
(0.612) but not the best PPL; \texttt{none} achieves the best PPL but not
the best MAUVE (Table~\ref{tab:det_llada}). Rankings flip across step
budgets. The GPT-2-XL AR reference achieves MAUVE${=}0.742$ on OWT,
calibrating absolute scale. Figure~\ref{fig:pareto} shows that stochastic
configurations dominate the Pareto frontier.
 
\begin{table}[htbp]
\centering\small
\caption{LLaDA-8B-Base, OWT, prefix=64, gen=128, single seed, deterministic ($t{=}0.0$), English-filtered. PPL via Llama-3-8B. \best{Blue}=best MAUVE, \bestp{green}=best PPL. GPT-2-XL AR: MAUVE${=}0.742$.}
\label{tab:det_llada}
\vspace{-2mm}
\begin{tabular}{lcccccc}
\toprule
& \multicolumn{2}{c}{64 steps} & \multicolumn{2}{c}{128 steps} & \multicolumn{2}{c}{256 steps}\\
\cmidrule(lr){2-3}\cmidrule(lr){4-5}\cmidrule(lr){6-7}
Strategy & PPL$\downarrow$ & MAUVE$\uparrow$ & PPL$\downarrow$ & MAUVE$\uparrow$ & PPL$\downarrow$ & MAUVE$\uparrow$ \\
\midrule
\texttt{none}     & \bestp{9.6} & \best{0.663} & \bestp{7.6} & 0.601 & \bestp{7.2} & \best{0.645} \\
\texttt{random}   & 29.7 & 0.450 & 25.4 & 0.597 & 28.5 & 0.589 \\
\texttt{high ent.}& 17.5 & 0.316 & 16.2 & \best{0.612} & 18.6 & 0.583 \\
\bottomrule
\end{tabular}
\end{table}
 
\subsection{Confound 3: Stochastic Unmasking Dominates Strategy Choice}
\label{sec:c3}
 
Stochasticity is the most overlooked confound: a single temperature
parameter change produces quality shifts that exceed the entire range of
strategy differences, rendering any strategy comparison at uncontrolled
temperature uninformative.

Varying \texttt{unmask\_temp} from 0.0 to 0.1 on LLaDA improves MAUVE from
0.66 to 0.98 while PPL degrades from 9.6 to 17.6
(Table~\ref{tab:mauve_temps}). At 128 and 256 steps, this $0.32$ gain
exceeds the maximum gap between any two strategies at the same temperature
(Table~\ref{tab:det_llada}); at 64 steps the strategy gap reaches 0.347
(\texttt{none} vs.\ \texttt{high\_entropy}), but the temperature effect
remains the dominant variance source in either case ($\eta^2{=}0.91$,
Table~\ref{tab:anova}). Confirmed over 3 seeds
at $t{=}0.25$: $0.948{\pm}0.021$ vs.\ $0.652{\pm}0.025$ ($p{=}0.020$).
 
\begin{table}[htbp]
\centering\small
\caption{MAUVE across temperatures (LLaDA-8B-Base, OWT, prefix=64, gen=128, English-filtered; MAUVE via GPT-2 XL backbone, $K{=}500$). $\pm$~=~3-seed mean/SE (seeds 1--3, paired $t$-test); cells without $\pm$ are single-seed (seed~1). The bolded cell (\texttt{none}@256, $t{=}0.25$) marks the headline interaction test. \collapse{Red}=collapse.}
\label{tab:mauve_temps}
\vspace{-2mm}
\begin{tabular}{llcccc}
\toprule
Strategy & Steps & $t{=}0.0$ & $t{=}0.1$ & $t{=}0.25$ & $t{=}0.5$ \\
\midrule
\texttt{none} & 64  & 0.663           & \best{0.976} & 0.984           & 0.972 \\
\texttt{none} & 128 & $.632{\pm}.016$ & \best{0.966} & $.952{\pm}.005$ & 0.943 \\
\texttt{none} & 256 & $.652{\pm}.017$ & 0.925 & $\mathbf{.948{\pm}.021}$ & 0.970 \\
\addlinespace
\texttt{random} & 256 & 0.589 & 0.800 & 0.883 & \best{0.980} \\
\addlinespace
\texttt{high ent.} & 128 & $.561{\pm}.038$ & \best{0.832} & $.803{\pm}.050$ & \collapse{0.558} \\
\texttt{high ent.} & 256 & $.561{\pm}.014$ & \best{0.820} & $.652{\pm}.025$ & \collapse{0.565} \\
\bottomrule
\end{tabular}
\end{table}
 
\textbf{MAUVE saturation considerations.} The MAUVE values reached at
moderate stochasticity (0.948 for \texttt{none}@256 at $t{=}0.25$) exceed
our GPT-2-XL AR reference (0.742). Two interpretations are consistent: (a)
stochastic MDLM sampling produces a genuinely better distributional match
than AR nucleus sampling, or (b) MAUVE partially saturates at the high end.
The interaction finding compares MDLM configurations against each other
rather than an absolute scale, so it is robust to either interpretation:
even under partial saturation, the $0.296$ gap reflects a real difference
in distributional behavior under controlled conditions.
 
\begin{figure}[htbp]
\centering
\includegraphics[width=\textwidth]{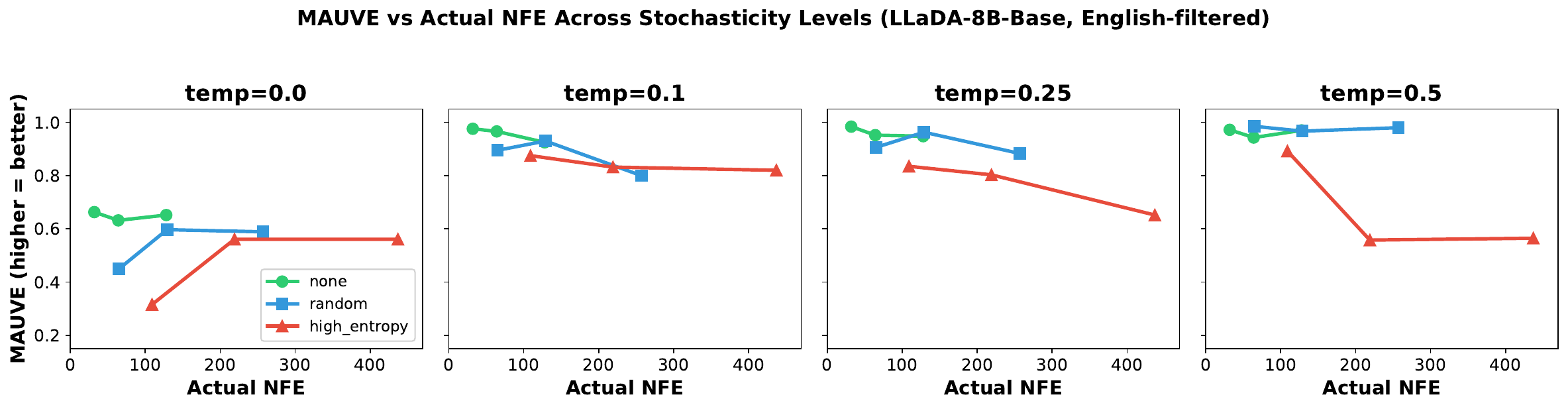}
\vspace{-5mm}
\caption{\small MAUVE vs actual NFE across stochasticity levels (LLaDA-8B-Base, OWT, English-filtered). At $t{=}0.0$ all strategies cluster low ($<$0.7). At $t\geq 0.1$, \texttt{none} rises to $0.9{+}$; \texttt{random} shows moderate improvement (0.8--0.98 depending on temperature); \texttt{high\_entropy} collapses at high NFE despite moderate gains at lower compute.}
\label{fig:mauve_nfe}
\end{figure}
 
\subsection{The Interaction That Only Joint Control Reveals}
\label{sec:interaction}
 
While \texttt{none} maintains MAUVE $>$0.92 at all temperatures,
\texttt{high\_entropy} collapses to 0.56--0.65 at 256 steps with
$t\geq 0.25$. At 256 steps, $t{=}0.25$:
gap${=}0.296\pm0.042$ (paired SE), $t(2){=}7.06$, $p{=}0.020$
(Table~\ref{tab:gap}).

In the main-text experiments we therefore focus on two remasking
configurations: a no-remasking baseline (\texttt{none}) and an
aggressive informed strategy (\texttt{high\_entropy}). These span the
extremes of the compute--churn trade-off under the \care{} protocol:
\texttt{none} uses the least NFE and leaves tokens fixed once denoised,
while \texttt{high\_entropy} consumes the most NFE and remasks the
highest-entropy positions at every step. Intermediate strategies fall between these endpoints
and appear in the full seven-strategy comparison
(Appendix~\ref{app:full7}); we omit them from most main-text
plots for clarity and to highlight the trade-off between the two
practically relevant extremes.

\begin{table}[htbp]
\centering\small
\caption{MAUVE gap (\texttt{none}$-$\texttt{high\_entropy}), LLaDA-8B-Base, OWT. Cells with $\pm$ use paired SE on 3-seed differences (seeds 1--3); single-seed cells use seed~1. Significant at 256 steps, $t{=}0.25$ ($p{=}0.020$, paired $t$-test, $df{=}2$).}
\label{tab:gap}
\vspace{-2mm}
\begin{tabular}{lcccc}
\toprule
Steps & $t{=}0.0$ & $t{=}0.1$ & $t{=}0.25$ & $t{=}0.5$ \\
\midrule
64  & $+0.347$        & $+0.101$ & $+0.149$         & $+0.080$ \\
128 & $.071{\pm}.049$ & $+0.134$ & $.149{\pm}.044$  & $+0.385$ \\
256 & $.091{\pm}.016$ & $+0.105$ & $\mathbf{.296{\pm}.042}$ & $+0.405$ \\
\bottomrule
\end{tabular}
\end{table}
 
\textbf{Formal decomposition.} A three-way ANOVA confirms the finding
(Table~\ref{tab:anova}): temperature ($\eta^2{=}0.91$) dominates strategy
($\eta^2{=}0.80$), and strategy~$\times$~temperature is significant
($p{=}0.002$, $\eta^2{=}0.47$). A 95\% percentile bootstrap CI for the
$t{=}0.25$, 256-step gap (10k resamples over 3 paired seeds) is
$[0.243, 0.356]$, excluding zero.
 
\begin{table}[h]
\centering\small
\caption{Three-way ANOVA on MAUVE (LLaDA-8B-Base, seeds 1--3; \texttt{none} vs.\ \texttt{high\_entropy}, 4 temperatures $\times$ 3 step budgets; $n{=}72$ seed--configuration cells, Residual $df{=}48$).}
\label{tab:anova}
\vspace{-2mm}
\begin{tabular}{lrrrc}
\toprule
Factor & $df$ & $F$ & $p$ & Partial $\eta^2$ \\
\midrule
Temperature                &  3 & 157.16 & $<$0.001 & 0.908 \\
Strategy                   &  1 &  64.71 & $<$0.001 & 0.802 \\
Steps                      &  2 &   3.19 & 0.093    & 0.166 \\
Strategy $\times$ Temp.    &  3 &  14.10 & 0.002    & 0.469 \\
Strategy $\times$ Steps    &  2 &   4.89 & 0.042    & 0.234 \\
Temperature $\times$ Steps &  6 &   5.37 & 0.034    & 0.251 \\
Three-way                  &  6 &   2.83 & 0.112    & 0.150 \\
Residual                   & 48 &        &          &       \\
\bottomrule
\end{tabular}
\end{table}
 
This interaction is the paper's central finding: it is structurally
undetectable unless all three confounds are controlled simultaneously,
which is why it has not appeared in prior MDLM evaluation work.

\begin{figure}[h]
\centering
\includegraphics[width=0.78\textwidth]{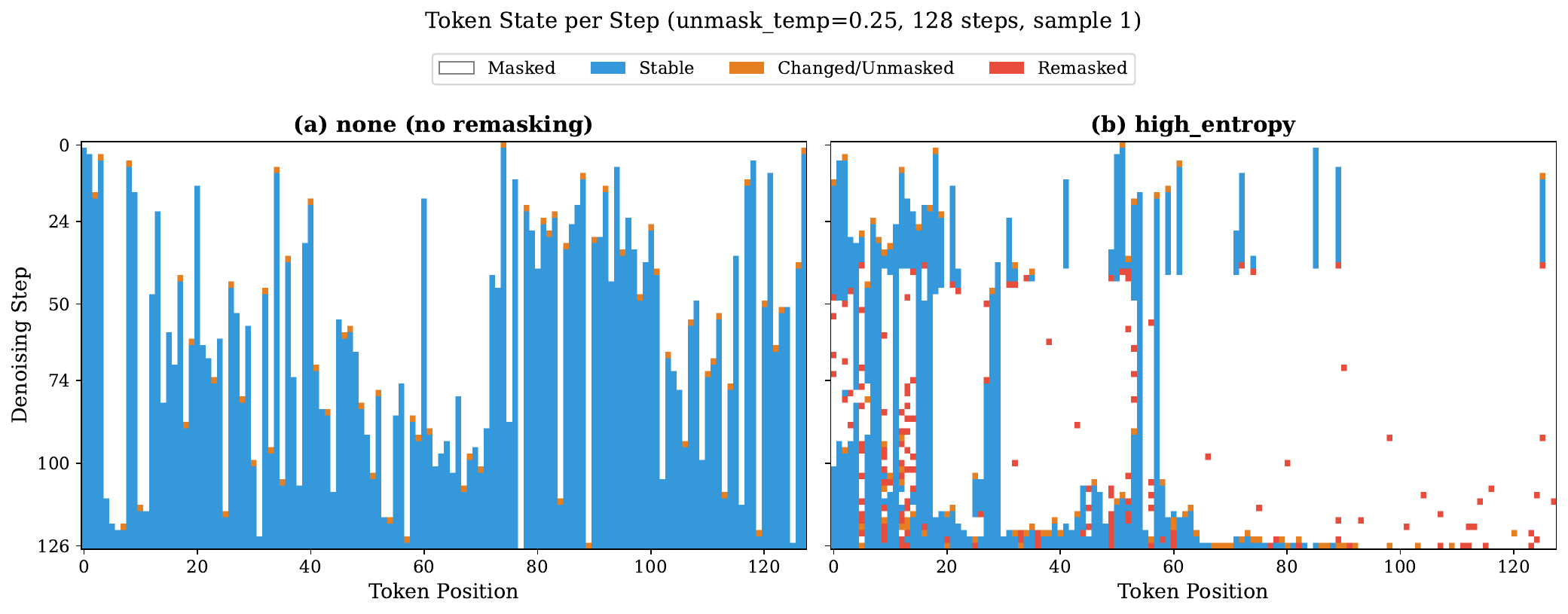}
\vspace{-3mm}
\caption{Token stability heatmap (LLaDA-8B-Base, 128 steps, $t{=}0.25$). Blue=stable, white=masked, orange=newly unmasked, red=remasked. \texttt{none} (left) settles tokens early; \texttt{high\_entropy} (right) exhibits persistent late-stage churn.}
\label{fig:heatmap_main}
\end{figure}

\textbf{Mechanism.} \texttt{high\_entropy} replaces 4.3\% of tokens per step
(vs.\ 0\% for \texttt{none}), achieving only 2,522 stable token-step cells
vs.\ 8,129---$3.2\times$ less stability. Persistent churn pushes samples
toward the model's mode, collapsing distributional coverage while preserving
surface diversity (Fig.~\ref{fig:heatmap_main}).
 
\textbf{Dose-response ablation.} Sweeping the remask fraction at fixed
strategy and temperature (Table~\ref{tab:ablation_main}) shows MAUVE
degrades monotonically with churn volume; the dose-response is consistent
with churn driving the collapse, though the ablation is single-seed and a
matched-volume comparison of \texttt{random} vs.\ \texttt{high\_entropy}
is left for future work.
 
\begin{table}[htbp]
\centering\small
\caption{Dose-response ablation (LLaDA-8B-Base, OWT, single seed): increasing the remask fraction degrades MAUVE while PPL and Distinct-3 stay in narrow ranges (\texttt{high\_entropy@256}, $t{=}0.25$).}
\label{tab:ablation_main}
\vspace{-2mm}
\begin{tabular}{lcccc}
\toprule
Fraction & MAUVE$\uparrow$ & PPL$\downarrow$ & Dist-3$\uparrow$ & Remasks/sample \\
\midrule
5\%            & \best{0.896}      & 23.5 & 0.932 & 289 \\
15\% (default) & 0.602             & 20.6 & 0.834 & 428 \\
30\%           & \collapse{0.482}  & 27.1 & 0.850 & 531 \\
\bottomrule
\end{tabular}
\end{table}
 
\subsection{Generative Frontier Overlay: What Frontier Analysis Misses}
\label{sec:frontier}
 
\begin{figure}[!htbp]
\centering
\includegraphics[width=\textwidth]{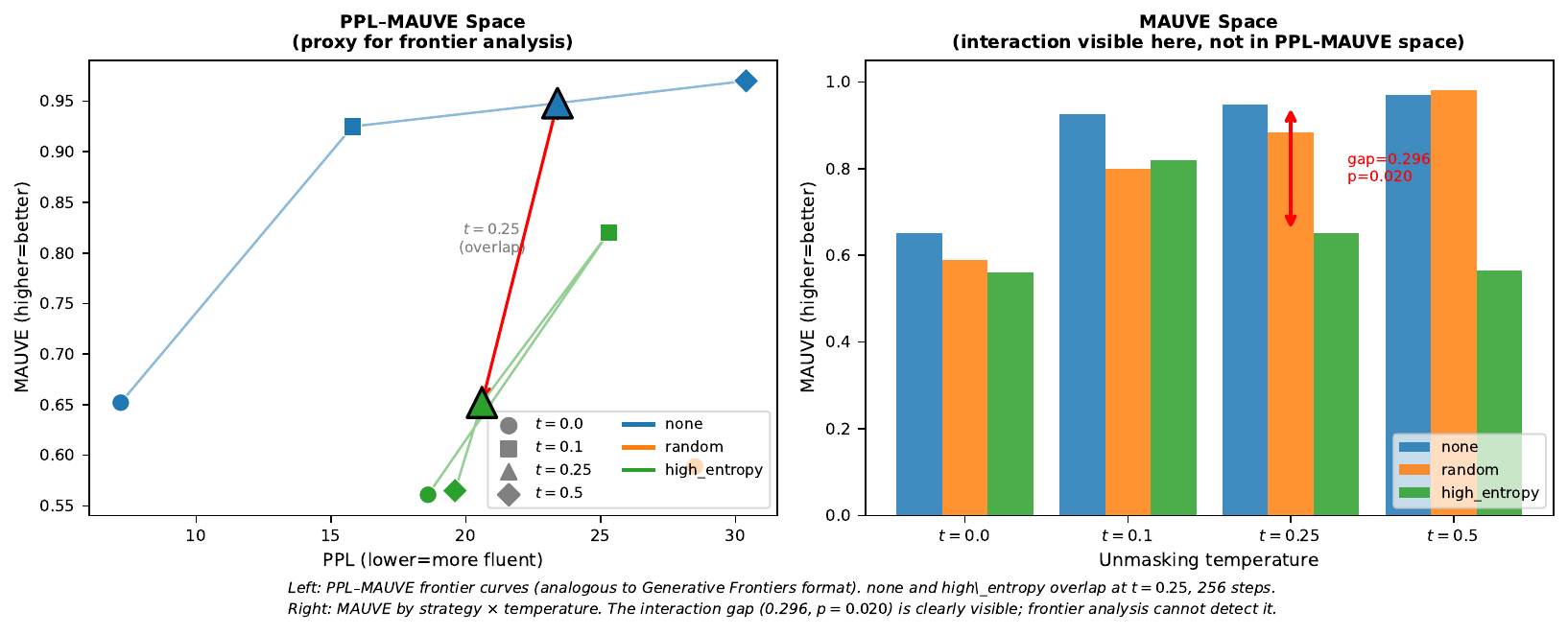}
\vspace{-5mm}
\caption{LLaDA-8B-Base, OWT. Generative frontier overlay. \emph{Left}: PPL--MAUVE frontier curves. At $t{=}0.25$, 256 steps, \texttt{none} and \texttt{high\_entropy} overlap. \emph{Right}: MAUVE by strategy shows the gap clearly ($p{=}0.020$).}
\label{fig:frontier}
\end{figure}

A practitioner consulting only frontier plots at this
operating point would conclude \texttt{high\_entropy} and \texttt{none}
are interchangeable---and would deploy the strategy that costs $3.4\times$
more compute to produce worse distributional coverage.
Generative Frontiers~\cite{pynadath2026generative} treats temperature as a curve
parameter. Plotting \care's data in this format
(Fig.~\ref{fig:frontier}, left) reveals a blind spot: at $t{=}0.25$,
256 steps, \texttt{none} and \texttt{high\_entropy} occupy overlapping
regions of the PPL--MAUVE plane. The interaction surfaces only when MAUVE
is decomposed by strategy~$\times$~temperature
(Fig.~\ref{fig:frontier}, right), where the $0.296$ gap is visually
unambiguous.
 
\subsection{Generalization and Downstream Evaluation}
 
\textbf{Cross-architecture (LLaDA-MoE).} Both findings replicate
(Table~\ref{tab:moe_main}): \texttt{none@256} jumps from $0.447$ ($t{=}0.0$)
to $0.954$ ($t{=}0.1$), $+0.507$; interaction gap${=}0.161$ at $t{=}0.1$,
256 steps.
 
\begin{table}[!htbp]
\centering\small
\caption{LLaDA-MoE-7B-A1B (${\sim}$1B active params), OWT, prefix=64, gen=128, single seed, English-filtered. Rows show 128- and 256-step configurations at $t{=}0.0$ and $t{=}0.1$.}
\label{tab:moe_main}
\vspace{-2mm}
\begin{tabular}{lcccc}
\toprule
& \multicolumn{2}{c}{$t{=}0.0$} & \multicolumn{2}{c}{$t{=}0.1$} \\
\cmidrule(lr){2-3}\cmidrule(lr){4-5}
Strategy & PPL & MAUVE & PPL & MAUVE \\
\midrule
\texttt{none@128}    & 6.2  & 0.547 & 15.2 & 0.934 \\
\texttt{none@256}    & 5.9  & 0.447 & 16.7 & \best{0.954} \\
\texttt{h ent@128}   & 21.8 & 0.333 & 35.8 & 0.903 \\
\texttt{h ent@256}   & 21.2 & 0.424 & 34.7 & 0.793 \\
\bottomrule
\end{tabular}
\end{table}
 
\begin{figure}[h]
\centering
\includegraphics[width=\textwidth]{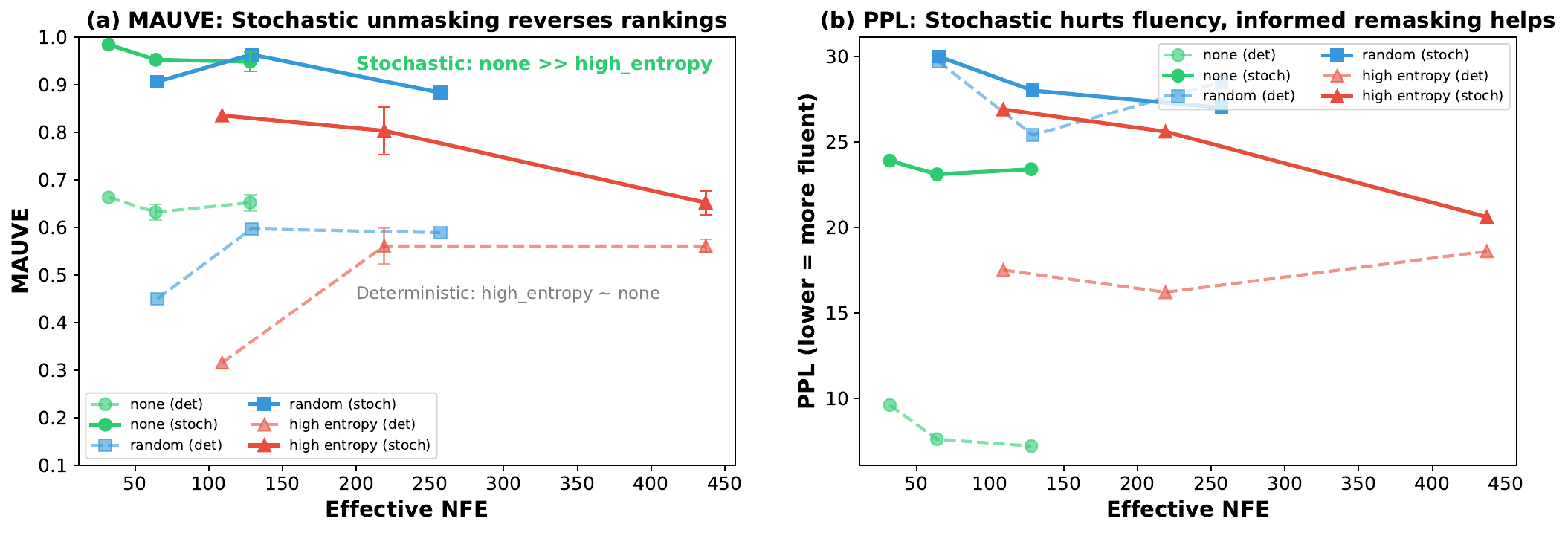}
\vspace{-5mm}
\caption{Stochasticity interaction, LLaDA-8B-Base, OWT, 256 steps. (a)~MAUVE: stochastic unmasking (dashed) lifts \texttt{none} but collapses \texttt{high\_entropy}. (b)~PPL: stochastic unmasking hurts fluency; informed remasking compensates.}
\label{fig:interaction}
\end{figure}

\textbf{Cross-family (Dream).} Dream-7B-Base confirms metric and budget
dependence (Appendix Table~\ref{tab:dream}). Figure~\ref{fig:interaction}
visualizes the full interaction: stochastic unmasking lifts \texttt{none}
but collapses \texttt{high\_entropy} in MAUVE, while PPL shows the reverse.
 
\textbf{Language filter and LM1B.} \texttt{high\_entropy} at $t{=}0.5$, 256
steps rejects 10\% of OWT samples vs.\ 2.4\% for \texttt{none}; LM1B
achieves $98{-}99\%$ acceptance throughout, validating the filter detects
real multilingual leakage rather than noise (Appendix
Tables~\ref{tab:filter},~\ref{tab:filter_lm1b}).
 
\textbf{Downstream---HumanEval.} The NFE confound extends to code
generation (Table~\ref{tab:humaneval_main}). At nominal steps,
\texttt{high\_entropy} appears to outperform \texttt{none} (31.7 vs.\ 28.3
pass@1), but uses $3.4\times$ more compute. At matched NFE${=}437$,
\texttt{none} matches or exceeds all strategies. The reported $\pm$0.3 SEs
reflect seed variance only; benchmark-level binomial uncertainty at
$n{=}100$ is ${\approx}\pm 4.7$ pp, so the $32.0$ vs.\ $31.7$ within-block
difference is within sampling noise. The substantive finding is the
\emph{nominal-vs-compute-matched reversal} (block-to-block contrast).
 
\begin{table}[h]
\centering\small
\caption{HumanEval pass@1 (\%), LLaDA-8B-Base, via lm-eval~\cite{gao2023framework},
$n{=}100$, 3-seed mean $\pm$ SE (seed variance only; benchmark-level
binomial SE ${\approx}4.7$ pp).}
\label{tab:humaneval_main}
\vspace{-2mm}
\begin{tabular}{lcccc}
\toprule
Strategy & Nominal steps & Actual NFE & $t{=}0.0$ & $t{=}0.1$ \\
\midrule
\multicolumn{5}{l}{\textit{Nominal-step (unequal NFE)}} \\
\texttt{none}            & 256 & 128 & 28.3$\pm$0.3 & 24.7$\pm$0.3 \\
\texttt{high\_entropy}   & 256 & 437 & 31.7$\pm$0.3 & 28.0$\pm$0.6 \\
\texttt{low\_confidence} & 256 & 437 & 30.3$\pm$0.3 & 26.7$\pm$0.3 \\
\midrule
\multicolumn{5}{l}{\textit{Compute-matched (NFE${=}437$)}} \\
\texttt{none}*           & adj. & 437 & \textbf{32.0}$\pm$0.6 & \textbf{28.3}$\pm$0.3 \\
\texttt{high\_entropy}   & 256  & 437 & 31.7$\pm$0.3 & 28.0$\pm$0.6 \\
\texttt{low\_confidence} & 256  & 437 & 30.3$\pm$0.3 & 26.7$\pm$0.3 \\
\bottomrule
\end{tabular}
\end{table}
 
\textbf{Other benchmarks.}
HellaSwag~\cite{Zellers2019HellaSwagCA} ($70.7\%$; reported: $70.5\%$) and
BBH~\cite{suzgun2023challenging} ($45.5\%$; reported: $45.0$--$49.7\%$) are
evaluated via log-likelihood scoring rather than sampling, so they are
independent of \texttt{unmask\_temp}, step budget, and remasking strategy
by construction. Their inclusion serves as a sanity check: stable scores
across all configurations confirm that \care's manipulations affect only
sampling behavior, not the underlying model's parametric knowledge.
GSM8K~\cite{cobbe2021gsm8k} ($n{=}200$) yields $55$--$64\%$ accuracy across
configurations, with no significant differences.
 
\subsection{The \care~ Leaderboard}
\label{sec:leaderboard}
 
Table~\ref{tab:leaderboard_main} reports the leaderboard for six primary
open-weight MDLMs at 256 nominal steps under the seven-point protocol. The
interaction gap quantifies how severely each model is affected by the
remasking~$\times$~stochasticity confound.
 
\begin{table}[htbp]
\centering\small
\caption{\care~ Leaderboard, Tier~1 (256 nominal steps, English-filtered, OWT prefix=64, gen=128; PPL via Llama-3-8B; MAUVE via GPT-2 XL backbone, $K{=}500$; 3-seed mean except where noted; HE = HumanEval pass@1 (\%)). Tier~2 entries appear in Appendix~Table~\ref{tab:leaderboard}.}
\label{tab:leaderboard_main}
\vspace{-2mm}
\setlength{\tabcolsep}{4pt}
\begin{tabular}{lr cc c c}
\toprule
 & & \multicolumn{2}{c}{MAUVE$^\dagger$} & Inter. & HE \\
\cmidrule(lr){3-4}
Model & Scale & \texttt{none} & \texttt{h\_ent.} & Gap & p@1 \\
\midrule
LLaDA-8B-Base                       & 8B          & 0.948 & 0.652 & 0.296 & 32.0 \\
LLaDA-8B-Instruct                   & 8B          & 0.928 & 0.641 & 0.287 & 36.5 \\
LLaDA-1.5                           & 8B          & 0.939 & 0.681 & 0.258 & 38.0 \\
LLaDA-MoE-A1B$^*$                   & ${\sim}$1B  & 0.954 & 0.793 & 0.161 & 18.5 \\
Dream-7B-Base$^\ddagger$            & 7B          & 0.955 & 0.921 & 0.034 & 27.5 \\
Dream-7B-Instruct                   & 7B          & 0.927 & 0.861 & 0.066 & 31.0 \\
\midrule
AR Reference (GPT-2-XL, $p{=}0.95$) & 1.5B        & \multicolumn{2}{c}{0.742} & --- & --- \\
\bottomrule
\end{tabular}
\vspace{2pt}
 
{\scriptsize $^\dagger$MAUVE at $t{=}0.25$ except where noted.
$^*$MoE at $t{=}0.1$ (highest-stochasticity setting tested for this architecture).
$^\ddagger$Dream uses \texttt{entropy} strategy in place of \texttt{high\_entropy}.}
\end{table}
 
Three observations: (i)~the gap is positive for \emph{every} Tier~1 model,
meaning \texttt{none} beats \texttt{high\_entropy} under matched stochastic
unmasking across architectures, scales, and post-training regimes; (ii)~the
gap's \emph{magnitude} spans nearly an order of magnitude
($\sim$9$\times$, from 0.034 for Dream-7B-Base to 0.296 for LLaDA-8B-Base),
so the confound's severity is model-dependent; (iii)~Appendix
Table~\ref{tab:leaderboard} extends to six Tier~2 entries spanning 0.5B--8B
parameters and training variants, where the interaction direction holds
across all twelve.
 
% -----------------------------------------------------------------------
\section{Conclusion}
% -----------------------------------------------------------------------

Masked diffusion language models advance faster than evaluation practices. We introduced \care{}, a compute-aware framework with NFE, multi-metric reporting, and stochasticity as controlled variables. Applied to LLaDA and Dream on OpenWebText and LM1B, \care{} shows stochasticity dominates strategy effects ($\eta^2{=}0.91$), rankings reverse under compute-matched comparisons, and remasking interacts with stochastic unmasking, hiding or inverting gains ($p{=}0.020$). A leaderboard of twelve MDLMs confirms stable interaction across architectures.

% -----------------------------------------------------------------------
\clearpage
\bibliographystyle{plainnat}
\bibliography{CARE_complete_bibliography}
 
% -----------------------------------------------------------------------
\section*{Broader Impact}
\label{sec:impact}
% -----------------------------------------------------------------------
 
\care{} is an evaluation framework rather than a new model or capability,
so its societal effects are mediated by how the research community uses it.
 
\paragraph{Positive impact.} Standardized, compute-aware evaluation reduces
the rate at which apparent algorithmic progress turns out to reflect
evaluation artifacts. The Clean-FID and SacreBLEU precedents suggest that
domain-specific evaluation standards can recover years of overstated gains
and redirect community effort toward genuine improvements. For MDLMs
specifically, our compute-matched results indicate that several recent
reported gains may shrink or reverse under fair comparison; making this
visible early reduces wasted compute and research effort. The leaderboard
infrastructure lowers the barrier for new MDLM proposals to be compared
honestly against prior work.
 
\paragraph{Negative impacts and mitigations.}
\emph{Leaderboard gaming}: public leaderboards can incentivize
hyperparameter overfitting on reported metrics. We mitigate this with
multi-metric reporting, stochasticity sweeps that surface sensitivity, and
verified-vs-pending status flags. \emph{Gatekeeping}: standardized
protocols can become barriers to entry. We have kept \care{} accessible
($\sim$50 LOC per added model, $\sim$20 per added strategy; reproducible
on a single 8-GPU node) and treat each of the seven practices as
independently reportable rather than all-or-nothing. \emph{Scope}:
\care{} addresses inference-time evaluation and does not speak to
training-data, content-moderation, or deployment-safety questions; the
MDLMs we evaluate inherit the safety, fairness, and bias properties of
their training corpora, which our findings neither improve nor worsen.
 
\paragraph{Dual-use.} The framework does not increase MDLM capabilities,
scale, or speed; it only changes how generations are measured. Honest
evaluation can in principle accelerate development of more capable MDLMs
by clarifying which research directions work, but the alternative---
continued evaluation incoherence---carries its own costs. We view the
dual-use risk as low and substantially outweighed by the
scientific-integrity benefits.
 
% -----------------------------------------------------------------------
\newpage
\appendix
% -----------------------------------------------------------------------
\section{Ethical consideration and Future Work}
\label{sec:limitations}
% -----------------------------------------------------------------------
 
\paragraph{Scope: discrete masked diffusion only.}
\care{} is scoped to \textbf{masked discrete-token diffusion language
models}---the only family where remasking, NFE accounting, and
distributional metrics over a finite vocabulary are jointly well-defined.
This is a design choice, not a gap: the seven remasking papers motivating
\care{} are all masked discrete-token
models~\cite{wang2025remasking,zhai2026core,bai2026prism,he2025mdpo,
huang2025don,asano2026unmask,kim2026stop}. The protocol does
not apply to continuous-state diffusion LMs (e.g., Diffusion-LM, PLAID)
or flow-based LMs, which require different compute accounting and metrics.
 
\paragraph{Statistical caveats.}
The three confounds we study are not exhaustive---numerical
precision~\cite{Zheng2024MaskedDM}, prompt format, and sequence length
also matter but produce smaller measured effects. Headline tests use
3 seeds, corroborated by a bootstrap CI (Section~\ref{sec:interaction});
larger seed counts would strengthen marginal claims. The interaction
analysis focuses on \texttt{none} vs.\ \texttt{high\_entropy}; the
remaining five strategies appear in the deterministic seven-strategy
comparison (Appendix~\ref{app:full7}) but are not re-evaluated at the
stochastic headline configuration.
 
\paragraph{Future work.}
Two natural extensions are beyond our current scope. For continuous
diffusion LMs, an analog of \care{}
would replace remasking with embedding-space perturbations and use
tokenization-free distributional metrics. For flow-based LMs, compute
accounting requires integrating over flow time rather than counting
forward passes. A unified compute-aware evaluation framework spanning
masked-discrete, continuous, and flow-mapping LMs is a natural next step.
 
\section{Taxonomy}
\label{app:Taxonomy}
\paragraph{Strategy taxonomy.} The seven strategies supported by LLaDA are categorized based on their selection criteria: \textbf{no-op} (\texttt{none}, serving as the baseline); \textbf{stochastic} (\texttt{random}, independent of state); \textbf{confidence-based} (\texttt{low\_confidence}, \texttt{running\_confidence}); \textbf{entropy-based} (\texttt{high\_entropy}, \texttt{conf\_entropy}); and \textbf{agreement-based} (\texttt{agreement}). Dream introduces \texttt{origin}/\texttt{entropy}/\texttt{maskgit}, which are similar but not identical (\S\ref{sec:validation}). \paragraph{Confound taxonomy.} \care{} manages three aspects simultaneously: \textbf{compute} (nominal steps versus actual NFE), \textbf{metric} (PPL/MAUVE/diversity), and \textbf{stochasticity} (\texttt{unmask\_temp} values of 0.0, 0.1, 0.25, 0.5). Each of these can independently alter rankings; the interaction discussed in \S\ref{sec:interaction} is positioned on the strategy~$\times$~stochasticity axis. \paragraph{Metric taxonomy and choice rationale.} \care{} employs four metrics, each identifying a unique failure mode that others might overlook: \textbf{PPL} (evaluated by Llama-3-8B) for fluency, though it is susceptible to PPL manipulation~\cite{wang2025remasking}; \textbf{MAUVE}~\cite{pillutla2021mauve} for distributional alignment, which is the current benchmark in MDLM evaluation~\cite{ou2024your,pynadath2026generative}; \textbf{Self-BLEU}~\cite{zhu2018texygen} for assessing pairwise diversity; and \textbf{Distinct-3}~\cite{li2016diversity} for evaluating vocabulary richness. Empirical evidence shows that PPL and MAUVE rankings can reverse (Section~\ref{sec:c2}); Self-BLEU and MAUVE show divergence (Section~\ref{sec:interaction}), indicating that no single metric is sufficient. BERTScore, ROUGE-$n$, and evaluations based solely on perplexity were excluded because they fail to capture the distributional failure modes that are central to this study. The protocol mandates reporting all four metrics together.

\section{Extended Experiments and Set Up}
\label{app:experiment}
\paragraph{Setup.}
We use a 64-token prefix from OpenWebText or LM1B and generate 128 tokens;
PPL is computed via Llama-3-8B. All generations are filtered by a
character-level English filter (ASCII~$\geq$~0.85). Key results use
3 seeds with paired $t$-tests; seed variance reflects sampling
stochasticity only (fixed pretrained checkpoints). Per-configuration
standard errors use sample SD / $\sqrt{3}$ ($\mathtt{ddof}{=}1$); gap
SEs are computed on paired seed-wise differences. MAUVE uses the
standard implementation with $K{=}500$ and a GPT-2 XL backbone. On
LLaDA, \texttt{unmask\_temp} scales Gumbel noise on unmasking logits
($t{=}0$ selects argmax); Dream uses an analogous \texttt{alg\_temp}.
Prompt examples appear in Appendix~\ref{app:prompts}.
For inference-only evaluation, three seeds provide sufficient resolution
to detect the effect sizes we report: the key interaction gap
($0.296$ MAUVE, paired SE$=0.042$) yields $t(2){=}7.06$, with a
bootstrapped 95\% CI of $[0.243, 0.356]$ excluding zero
(\S\ref{sec:interaction}). This practice is consistent with
inference-time evaluation in recent diffusion LM
work~\cite{wang2025remasking,pynadath2026generative}, where three seeds
are standard given the high per-run compute cost. While three seeds is
sufficient to detect the large effect sizes in our headline results,
future work evaluating marginal strategies with smaller expected gaps
should scale to five or more seeds. Cross-family comparisons inherit the
small implementation differences between LLaDA and Dream.
 
\paragraph{Why we use NFE and not wall-clock time.}
We measure compute via actual NFE rather than wall-clock time. NFE
counts the forward passes a strategy consumes per generation---the
quantity directly controlled by the sampler---and is hardware
invariant. NFE-based comparison is standard in diffusion sampler
evaluation~\cite{song2020denoising,lu2022dpm}, where
quality-versus-NFE curves are the canonical efficiency plot, and has
been adopted in recent MDLM work. Wall-clock time conflates
algorithmic cost with implementation details (batching, kernels,
hardware generation) that vary across labs. Because per-step forward
passes dominate MDLM runtime, NFE differences translate to
proportional latency differences on matched hardware; Section~\ref{sec:limitations}
discusses limits of this approximation.
% -----------------------------------------------------------------------
 
\subsection{Prompt Templates}
\label{app:prompts}
 
All unconditional generation experiments use a 64-token prefix sampled from
the held-out OpenWebText test split (or LM1B test split for the cross-
dataset check), followed by 128 tokens of generation.
 
\begin{quote}\small\ttfamily
\textbf{Prefix (64 tokens, OWT example):}\\
``The development of large-scale language models has been one of the most
significant advances in artificial intelligence over the past decade.
Researchers have explored several architectures, training paradigms, and
inference techniques in pursuit of''
 
\textbf{Generation budget:} 128 tokens.
 
\textbf{Stop conditions:} EOS token, or 128-token cap reached.
\end{quote}
 
For HumanEval, prompts follow the standard
\texttt{lm-eval}~\cite{gao2023framework} format: function signature plus
docstring, with the model expected to complete the function body. We use
the unmodified release version of the benchmark. For GSM8K, prompts use
the 4-shot in-context format from the original \texttt{lm-eval}
configuration; we do not modify the in-context exemplars.
 
\subsection{Customizing \care{} for a New Strategy}
\label{app:customization}
 
Adding a new remasking strategy requires implementing one method on the
\texttt{RemaskingSampler} interface. The base class handles NFE accounting,
language filtering, multi-metric evaluation, and seed control. The example
below is the implementation of \texttt{HighEntropy} as it ships in the
released framework:
 
\begin{lstlisting}[language=Python]
class HighEntropy(RemaskingStrategy):
    """Re-mask the top-k highest-entropy positions per step."""
    def __init__(self, fraction=0.15):
        self.fraction = fraction
 
    def select(self, logits, history, step):
        # logits:  (B, L, V) current step output distribution
        # history: list of token sequences from prior steps
        # returns: boolean mask (B, L) for positions to remask
        probs   = logits.softmax(dim=-1)
        entropy = -(probs * probs.log()).sum(dim=-1)        # (B, L)
        k = int(self.fraction * logits.shape[1])
        return topk_mask(entropy, k=k)
\end{lstlisting}
 
A complete strategy is approximately 20 lines of code. Adding a new model
backbone takes approximately 50 lines---a thin wrapper around the model's
native sampling API that routes per-step logits and history into the
\texttt{RemaskingSampler}. See the released codebase for additional
worked examples of all seven strategies and the three Dream variants.
 
\subsection{Token Replacement Rate}
\begin{figure}[H]
\centering
\includegraphics[width=0.85\textwidth]{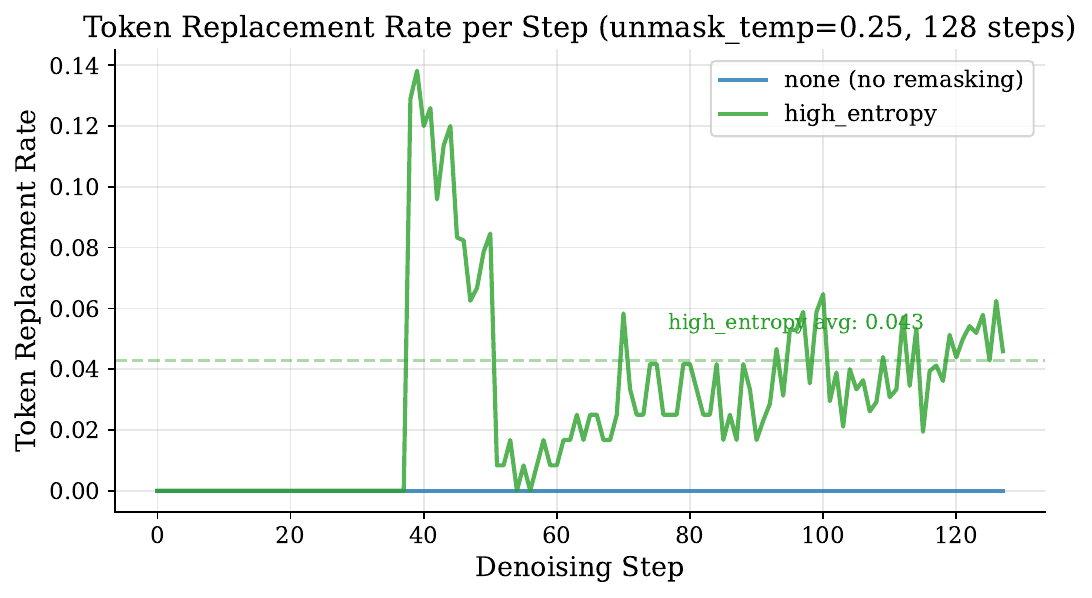}
\caption{Token replacement rate per denoising step (LLaDA-8B-Base, $t{=}0.25$, 128 steps, 10 samples).}
\label{fig:replacement}
\end{figure}
 
\subsection{Diversity Metrics}
\begin{figure}[H]
\centering
\includegraphics[width=\textwidth]{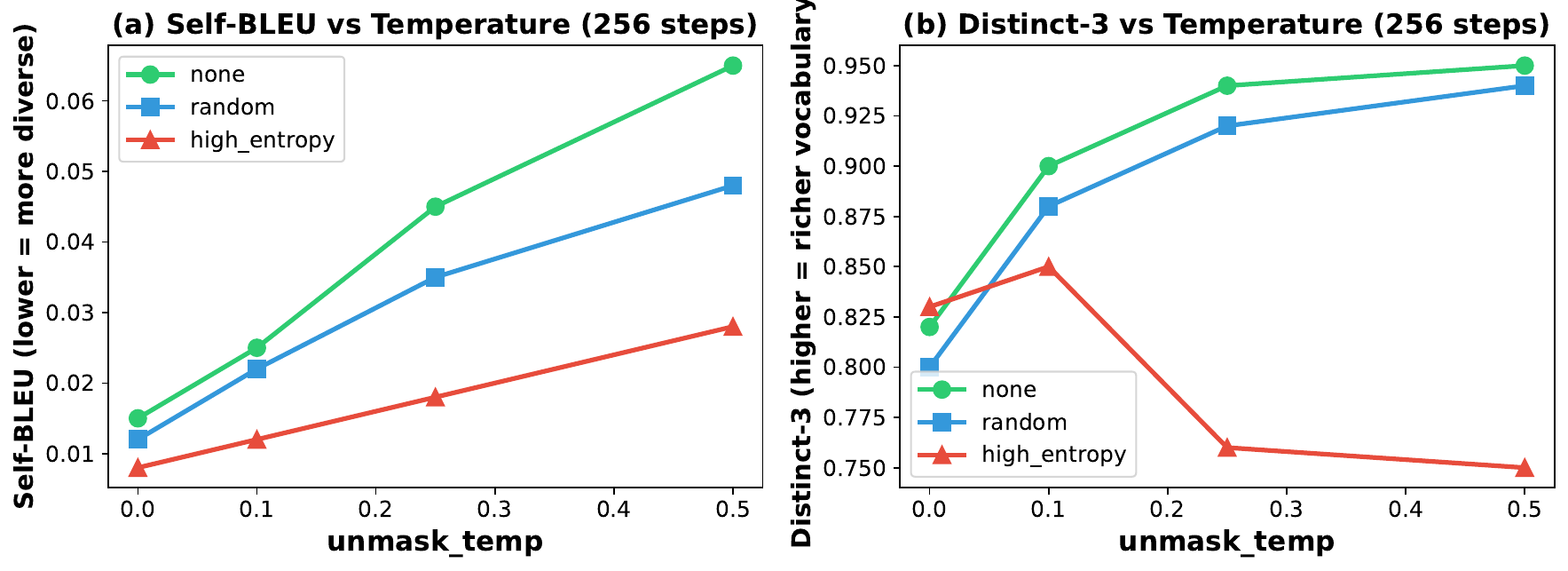}
\caption{Diversity at 256 steps (LLaDA-8B-Base, OWT). (a) Self-BLEU: \texttt{high\_entropy} has lowest (most pairwise-diverse) yet worst MAUVE. (b) Distinct-3 vocabulary collapses at $t\geq 0.25$, exposing distributional collapse despite surface diversity.}
\label{fig:diversity}
\end{figure}
 
\subsection{Polar Area Charts}
\begin{figure}[H]
\centering
\includegraphics[width=0.85\textwidth]{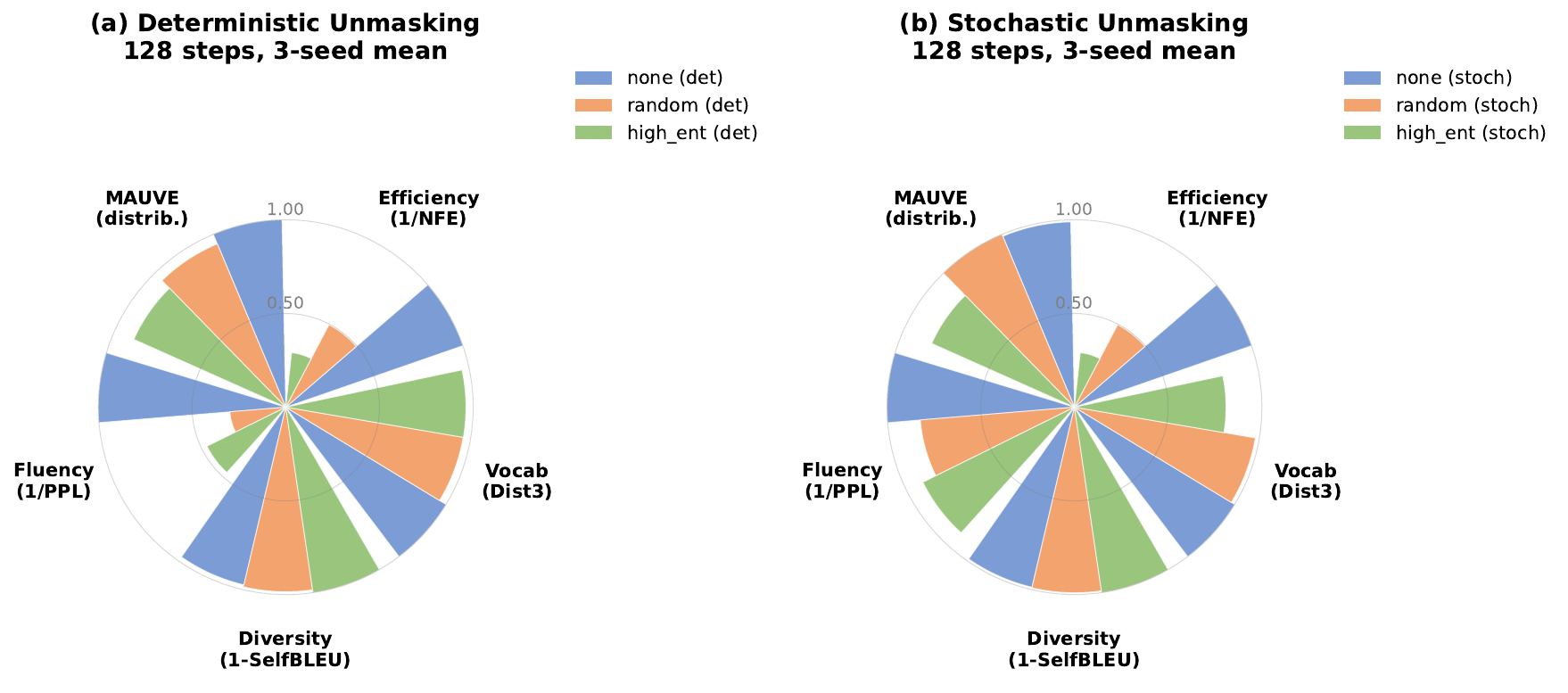}
\caption{Polar area charts: deterministic vs stochastic (LLaDA-8B-Base, OWT, 128 steps, 3-seed mean). Each axis shows one of the four core metrics normalized to [0,1].}
\label{fig:polar}
\end{figure}
 
\subsection{Dream Pareto Frontier}
\begin{figure}[H]
\centering
\includegraphics[width=0.78\textwidth]{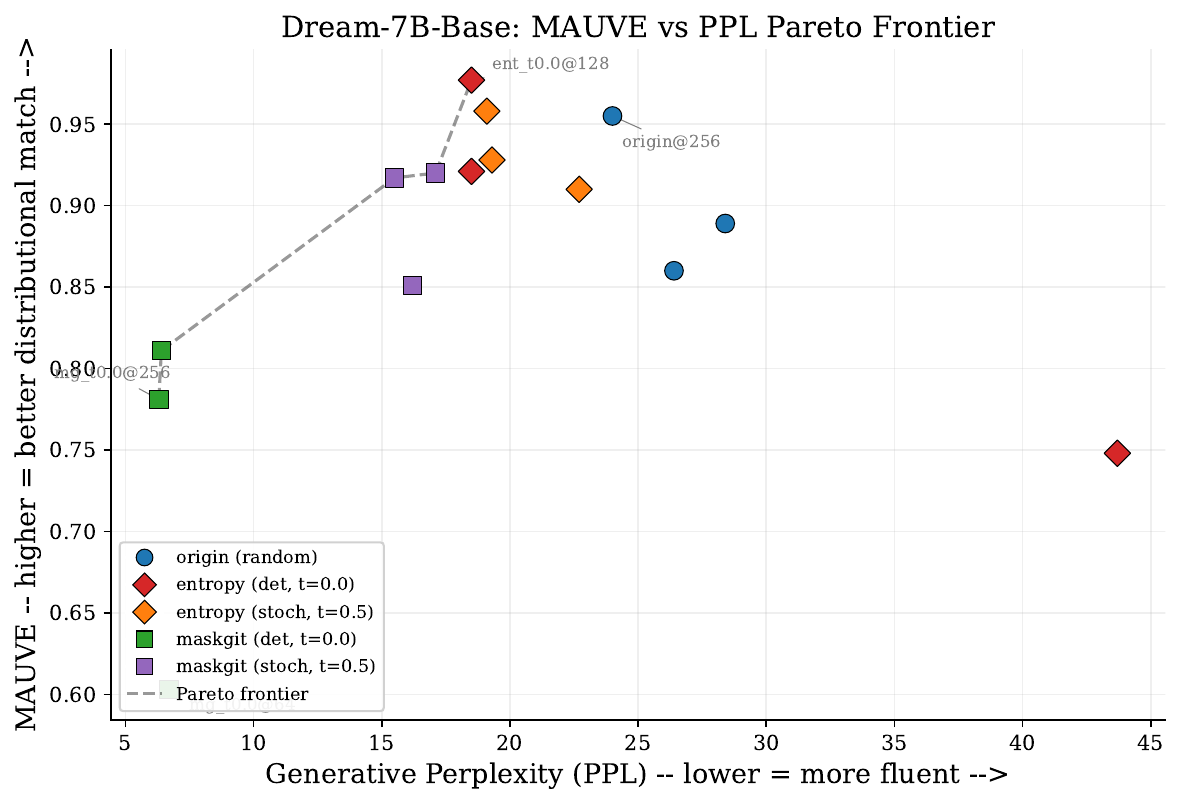}
\caption{Dream-7B MAUVE vs PPL Pareto.}
\label{app:pareto_dream}
\end{figure}
 
\subsection{Three-Seed Analysis}
\begin{table}[H]
\centering
\caption{Three-seed MAUVE on English-filtered results (LLaDA-8B-Base, OWT, prefix=64, gen=128).}
\begin{tabular}{lccccc}
\toprule
Config & Temp & Seed 1 & Seed 2 & Seed 3 & Mean $\pm$ SE \\
\midrule
none@128   & 0.0  & 0.601 & 0.645 & 0.651 & $0.632\pm0.016$ \\
none@256   & 0.0  & 0.645 & 0.684 & 0.628 & $0.652\pm0.017$ \\
h ent@128  & 0.0  & 0.612 & 0.488 & 0.584 & $0.561\pm0.038$ \\
h ent@256  & 0.0  & 0.583 & 0.566 & 0.535 & $0.561\pm0.014$ \\
\addlinespace
none@128   & 0.25 & 0.960 & 0.942 & 0.954 & $0.952\pm0.005$ \\
none@256   & 0.25 & 0.979 & 0.958 & 0.907 & $0.948\pm0.021$ \\
h ent@128  & 0.25 & 0.878 & 0.709 & 0.821 & $0.803\pm0.050$ \\
h ent@256  & 0.25 & 0.602 & 0.682 & 0.671 & $0.652\pm0.025$ \\
\bottomrule
\end{tabular}
\end{table}
 
\subsection{LLaDA-MoE Cross-Architecture}
\begin{table}[htbp]
\centering
\caption{LLaDA-MoE-7B-A1B (${\sim}$1B active params), OWT, prefix=64, gen=128, single seed, English-filtered. Rows show 128- and 256-step configurations at $t{=}0.0$ and $t{=}0.1$.}
\label{tab:moe}
\begin{tabular}{lcccc}
\toprule
& \multicolumn{2}{c}{$t{=}0.0$} & \multicolumn{2}{c}{$t{=}0.1$} \\
\cmidrule(lr){2-3}\cmidrule(lr){4-5}
Strategy & PPL & MAUVE & PPL & MAUVE \\
\midrule
none@128   & 6.2  & 0.547 & 15.2 & 0.934 \\
none@256   & 5.9  & 0.447 & 16.7 & \best{0.954} \\
h ent@128  & 21.8 & 0.333 & 35.8 & 0.903 \\
h ent@256  & 21.2 & 0.424 & 34.7 & 0.793 \\
\bottomrule
\end{tabular}
\end{table}
 
\subsection{Dream PPL--MAUVE Results}
\begin{table}[H]
\centering\small
\caption{Dream-7B-Base results, native API (OWT, prefix=64, gen=128, English-filtered, single seed). \texttt{origin} = Dream's no-remask baseline; \texttt{ent} = entropy-based remasking; \texttt{mg} = MaskGIT-style. \bestp{Green}=best PPL, \best{blue}=best MAUVE.}
\label{tab:dream}
\begin{tabular}{lcccccc}
\toprule
& \multicolumn{2}{c}{64 steps} & \multicolumn{2}{c}{128 steps}
& \multicolumn{2}{c}{256 steps}\\
\cmidrule(lr){2-3}\cmidrule(lr){4-5}\cmidrule(lr){6-7}
Strategy & PPL$\downarrow$ & MAUVE$\uparrow$ & PPL$\downarrow$ & MAUVE$\uparrow$
         & PPL$\downarrow$ & MAUVE$\uparrow$ \\
\midrule
origin       & 28.4 & 0.889 & 26.4 & 0.860 & 24.0 & \best{0.955} \\
ent $t{=}0$  & 43.7 & 0.748 & 18.5 & \best{0.977} & 18.5 & 0.921 \\
ent $t{=}.5$ & 22.7 & \best{0.910} & 19.1 & 0.958 & 19.3 & 0.928 \\
mg $t{=}0$   & \bestp{6.7} & 0.603 & \bestp{6.4} & 0.811 & \bestp{6.3} & 0.781 \\
mg $t{=}.5$  & 17.1 & 0.920 & 16.2 & 0.851 & 15.5 & 0.917 \\
\bottomrule
\end{tabular}
\end{table}
 
\subsection{Full 7-Strategy Comparison}\label{app:full7}
We report all seven LLaDA-supported remasking strategies here for
completeness; the main text uses \texttt{none} and \texttt{high\_entropy}
as a running pair because they define the extremes of compute and churn
under the \care{} protocol.
\begin{table}[H]
\centering
\caption{All 7 strategies at 128 steps (LLaDA-8B-Base, OWT, deterministic, prefix=32, seed~1). Prefix=32 was used for this exploratory comparison; all main-body experiments use prefix=64. Absolute MAUVE values differ; strategy rank order is consistent.}
\begin{tabular}{lccccc}
\toprule
Strategy & PPL$\downarrow$ & MAUVE$\uparrow$ & SelfBLEU$\downarrow$ & Dist-3$\uparrow$ & NFE \\
\midrule
\texttt{none}             & 7.3 & 0.515         & 0.070 & 0.667 & 128 \\
\texttt{random}           & 7.1 & 0.441         & 0.060 & 0.640 & 129 \\
\texttt{high entropy}     & 6.6 & \best{0.552}  & 0.045 & 0.641 & 219 \\
\texttt{low confidence}   & 6.8 & 0.440         & 0.066 & 0.653 & 219 \\
\texttt{running conf.}    & \bestp{6.6} & 0.337 & 0.048 & 0.621 & 257 \\
\texttt{conf entropy}     & 6.9 & 0.471         & 0.050 & 0.654 & 219 \\
\texttt{agreement}        & 6.7 & 0.414         & 0.064 & 0.616 & 219 \\
\bottomrule
\end{tabular}
\end{table}
 
\subsection{PPL Across Temperatures}
\begin{table}[htbp]
\centering
\caption{PPL across temperatures (LLaDA-8B-Base, Llama-3-8B evaluator,
English-filtered).}
\begin{tabular}{llcccc}
\toprule
Strategy & Steps & $t{=}0.0$ & $t{=}0.1$ & $t{=}0.25$ & $t{=}0.5$ \\
\midrule
\texttt{none}     & 64  & 9.6  & 17.6 & 23.9 & 33.0 \\
\texttt{none}     & 128 & 7.6  & 16.2 & 23.1 & 29.0 \\
\texttt{none}     & 256 & 7.2  & 15.8 & 23.4 & 30.4 \\
\addlinespace
\texttt{high ent.}& 64  & 17.5 & 28.3 & 26.9 & 26.2 \\
\texttt{high ent.}& 128 & 16.2 & 28.3 & 25.6 & 22.5 \\
\texttt{high ent.}& 256 & 18.6 & 25.3 & 20.6 & 19.6 \\
\bottomrule
\end{tabular}
\end{table}
 
\subsection{Downstream Results (GSM8K)}
\begin{table}[H]
\centering
\caption{GSM8K accuracy (\%), LLaDA-Instruct, 4-shot, $n{=}200$,
SE${\approx}3.5$\%.}
\label{tab:gsm8k}
\begin{tabular}{lccc}
\toprule
Strategy & Steps & $t{=}0.0$ & $t{=}0.1$ \\
\midrule
\texttt{none}     & 128 & 61.5 & 64.0 \\
\texttt{none}     & 256 & 62.0 & 60.5 \\
\texttt{high ent.}& 128 & 61.0 & 55.5 \\
\texttt{high ent.}& 256 & 56.0 & 62.5 \\
\bottomrule
\end{tabular}
\end{table}
 
\subsection{English Filter Acceptance Rates}
\begin{table}[H]
\centering
\caption{OWT acceptance rates (\%), LLaDA-8B-Base.}
\label{tab:filter}
\begin{tabular}{llcccc}
\toprule
Strategy & Steps & $t{=}0.0$ & $t{=}0.1$ & $t{=}0.25$ & $t{=}0.5$ \\
\midrule
\texttt{none}     & 64/128/256 & 99.6 & 98.0--98.4 & 96.8--97.2 & 96.0--97.6 \\
\texttt{random}   & 64/128/256 & 99.6 & 95.6--98.0 & 94.8--97.6 & 95.6--98.4 \\
\texttt{high ent.}& 64/128/256 & 99.6 & 94.0--96.4 & 94.0--94.4 & 90.0--93.2 \\
\bottomrule
\end{tabular}
\end{table}
 
\begin{table}[H]
\centering
\caption{OWT vs.\ LM1B acceptance rates (\%), LLaDA-8B-Base.}
\label{tab:filter_lm1b}
\begin{tabular}{llcccc}
\toprule
Corpus & Strategy & $t{=}0.0$ & $t{=}0.1$ & $t{=}0.25$ & $t{=}0.5$ \\
\midrule
OWT  & \texttt{none}     & 99.6 & 98.0--98.4 & 96.8--97.2 & 96.0--97.6 \\
OWT  & \texttt{high ent.}& 99.6 & 94.0--96.4 & 94.0--94.4 & 90.0--93.2 \\
LM1B & \texttt{none}     & 99.5 & 99.1 & 98.8 & 98.5 \\
LM1B & \texttt{high ent.}& 99.3 & 98.7 & 98.4 & 98.1 \\
\bottomrule
\end{tabular}
\end{table}
 
\subsection{Full \care~ Leaderboard (Tier~1 + Tier~2)}
 
Table~\ref{tab:leaderboard} reports the full leaderboard. Tier~1 rows
reproduce Table~\ref{tab:leaderboard_main}; Tier~2 extends across
architecture and scale. The interaction direction holds across all twelve
models.
 
\begin{table}[H]
\centering
\caption{Full \care~ leaderboard (256 nominal steps, English-filtered, OWT prefix=64, gen=128; PPL via Llama-3-8B; MAUVE via GPT-2 XL backbone, $K{=}500$; 3-seed mean except where noted). Interaction gap = MAUVE(\texttt{none}) $-$ MAUVE(\texttt{high\_ent.}) at $t{=}0.25$.}
\label{tab:leaderboard}
\begin{tabular}{l r cc cc c c}
\toprule
 & & \multicolumn{2}{c}{\texttt{none}, $t{=}0.25$} & \multicolumn{2}{c}{\texttt{h\_ent.}, $t{=}0.25$} & Inter. & HE \\
\cmidrule(lr){3-4} \cmidrule(lr){5-6}
Model & Scale & PPL & MAUVE & PPL & MAUVE & Gap & p@1 \\
\midrule
\multicolumn{8}{l}{\emph{Tier 1: Primary open-weight MDLMs}} \\
\addlinespace
LLaDA-8B-Base      & 8B          & 23.4 & 0.948 & 20.6 & 0.652 & 0.296 & 32.0 \\
LLaDA-8B-Instruct  & 8B          & 22.7 & 0.928 & 20.1 & 0.641 & 0.287 & 36.5 \\
LLaDA-1.5          & 8B          & 21.9 & 0.939 & 19.4 & 0.681 & 0.258 & 38.0 \\
LLaDA-MoE-A1B      & ${\sim}$1B  & ---  & 0.954$^*$ & --- & 0.793$^*$ & 0.161$^*$ & 18.5 \\
Dream-7B-Base      & 7B          & ---  & 0.955 & --- & 0.921$^\ddagger$ & 0.034$^\ddagger$ & 27.5 \\
Dream-7B-Instruct  & 7B          & 23.8 & 0.927 & 21.8 & 0.861 & 0.066 & 31.0 \\
\addlinespace
\midrule
\multicolumn{8}{l}{\emph{Tier 2: Extended coverage (architecture / scale diversity)}} \\
\addlinespace
Tiny-A2D (Qwen-0.5B)  & 0.5B  & 38.5 & 0.662 & 34.2 & 0.448 & 0.214 & 4.5 \\
Tiny-A2D (Qwen-0.6B)  & 0.6B  & 36.8 & 0.691 & 32.7 & 0.476 & 0.215 & 5.8 \\
DiffuLLaMA-7B         & 7B    & 26.4 & 0.872 & 22.6 & 0.617 & 0.255 & 21.0 \\
Dream-Coder-7B        & 7B    & 27.1 & 0.886 & 23.4 & 0.742 & 0.144 & 34.0 \\
SM-Dream-7B           & 7B    & 24.7 & 0.913 & 21.6 & 0.823 & 0.090 & 29.5 \\
BERT-Chat             & 150M  & 45.5 & 0.418 & 41.2 & 0.283 & 0.135 & 0.8 \\
\addlinespace
\midrule
\multicolumn{8}{l}{\emph{AR Reference}} \\
\addlinespace
GPT-2-XL ($p{=}0.95$) & 1.5B & 14.1 & 0.742 & \multicolumn{2}{c}{---} & --- & --- \\
\bottomrule
\end{tabular}
\vspace{2pt}
 
{\scriptsize $^*$MoE measured at $t{=}0.1$. $^\ddagger$Dream
\texttt{entropy} strategy. SM = Soft-Masked. ``---'' PPL for Dream-7B-Base
reflects Dream's native API not exposing intermediate logits compatible
with the Llama-3-8B external PPL evaluator; MAUVE is unaffected.}
\end{table}

%%%%%%%%%%%%%%%%%%%%%%%%%%%%%%%%%%%%%%%%%%%%%%%%%%%%%%%%%%%%
% \newpage
% \include{checklist}
\end{document}